\documentclass[10pt,twocolumn,letterpaper]{article}

\usepackage[pagenumbers]{cvpr} 
\usepackage[utf8]{inputenc}
\usepackage{graphicx}
\usepackage{subcaption}
\usepackage{multirow}
\usepackage{color}
\usepackage{xcolor}
\usepackage{amsmath}
\usepackage{amssymb}
\usepackage{pifont}
\usepackage{booktabs}
\usepackage{ulem}
\usepackage{comment}

\renewcommand{\emph}[1]{\textit{#1}}

\definecolor{darkgreen}{rgb}{0.0, 0.5, 0.0}

\newcommand{\cmark}{\ding{51}}%
\newcommand{\xmark}{\ding{55}}%

\newcommand{\latentspace}{\mathcal{S}}
\newcommand{\latentcode}{s}

\newcommand{\hidden}{h}

\newcommand{\pointcloud}{P}

\newcommand{\lstringspace}{\mathcal{L}}
\newcommand{\lstring}{l}

\newcommand{\node}{n}

\newcommand{\dimension}{dim}

\newcommand{\numberNodeType}{N}

\newcommand{\encoderfunction}{E}

\newcommand{\nodeencoderi}{\encoderfunction_{node,i}}
\newcommand{\siblingencoder}{\encoderfunction_{sib}}
\newcommand{\pcencoder}{\encoderfunction_{pc}}
\newcommand{\encoderfunctionpoints}{\encoderfunction_{points}}

\newcommand{\decoderfunction}{D}

\newcommand{\nodedecoderi}{\decoderfunction_{node,i}}
\newcommand{\siblingdecoder}{\decoderfunction_{sib}}
\newcommand{\pcdecoder}{\decoderfunction_{pc}}

\newcommand{\classifier}{C}
\newcommand{\classifierSplit}{\classifier_{split}}
\newcommand{\classifierNode}{\classifier_{node}}

\newcommand{\loss}{L}
\newcommand{\reconstructionloss}{\loss_{rec}}

\newcommand{\splitloss}{\loss_{split}}
\newcommand{\nodeloss}{\loss_{node}}
\newcommand{\totalloss}{\loss_{total}}
\newcommand{\pointcloudloss}{\loss_{points}}

\definecolor{cvprblue}{rgb}{0.21,0.49,0.74}
\usepackage[pagebackref,breaklinks,colorlinks,allcolors=cvprblue]{hyperref}

\def\paperID{} 
\def\confName{CVPR}
\def\confYear{2026}

\title{Learning  to Infer Parameterized Representations of Plants from 3D Scans}

\author{Samara Ghrer$^{1}$ \hspace{1.0cm} Christophe Godin$^{2}$ \hspace{1.0cm} Stefanie Wuhrer$^{1}$\\
$^{1}$ {Inria center at Université Grenoble Alpes, France}\\
$^{2}$ {Inria center at Lyon, France}\\
{\tt\small samara.ghrer@inria.fr, christophe.godin@inria.fr, stefanie.wuhrer@inria.fr}
}

\begin{document}
\twocolumn[{%
\renewcommand\twocolumn[1][]{#1}%
\maketitle

\begin{center}
    \includegraphics[width = 0.9\linewidth]{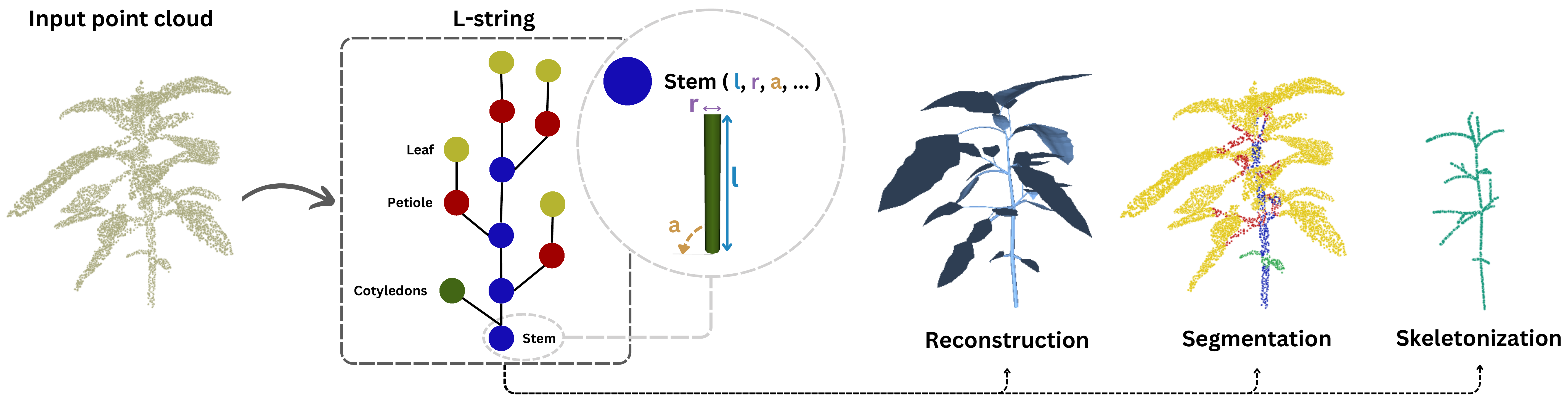}
    \captionof{figure}{Our method takes a 3D point cloud of a plant as input, and outputs a parameterized representation of the plant. This representation encodes the plant's branching structure and geometry along with semantic information such as organ type, and allows for multiple tasks including reconstruction, organ segmentation and skeleton extraction.}
    \label{fig:teaser}
\end{center} 
}]
\begin{abstract}
 Plants frequently contain numerous organs, organized in 3D branching systems defining the plant's architecture. Reconstructing the architecture of plants from unstructured observations is challenging because of self-occlusion and spatial proximity between organs, which are often thin structures. 
To achieve this challenging task, we propose an approach that allows to infer a parameterized representation of the plant's architecture from a given 3D scan of a plant. In addition to the plant's branching structure, this representation contains parametric information for each plant organ, and can therefore be used directly in a variety of tasks. In this data-driven approach, we train a recursive neural network with virtual plants generated using a procedural model. After training, the network allows to infer a parametric tree-like representation based on an input 3D point cloud. Our method is applicable to any plant that can be represented as binary axial tree. We quantitatively evaluate our approach on Chenopodium Album plants on reconstruction, segmentation and skeletonization, which are important problems in plant phenotyping. In addition to carrying out several tasks at once, our method achieves results on-par with strong baselines for each task. We apply our method, trained exclusively on synthetic data, to 3D scans and show that it generalizes well. Code available at~\href{https://gitlab.inria.fr/sghrer/3d-L-plants}{gitlab.inria.fr/sghrer/3d-L-plants}.
\end{abstract}    
\section{Introduction}
\label{sec:intro}

Plant phenotyping consists of quantifying how plant's genotypes grow in their environment and has important applications in crop management. To automate this task, it is necessary to infer high-level information of plants from observations. This problem, which has been studied for decades~\cite{phenotyping2015}, can be decomposed into a number of subtasks. High-level information includes the reconstruction of the 3D geometry of a plant, the segmentation of a plant into parts, and the extraction of a skeleton.
Observations are typically 2D images or 3D scans. Automatically extracting this information has applications in plant phenotyping,~\eg\cite{mirande2022graph, dobbs2023smart, ghahremani2021direct}, and to use reconstructed 3D plants in content generation and virtual reality~\cite{TreePartNet21}. 

Such an inference task is challenging as plants have complex structures due to their branching systems, which lead to strong self-occlusions and ambiguities. Existing works that analyze 2D or 3D plant observations therefore either focus on inverse modeling, where the goal is to find growth rules that allow to generate a given plant, or are task-specific. Inverse modeling is challenging and existing methods are limited to leafless branching structures~\cite{guo2020inverse,stava2014inverse}. Task-specific methods focus on 3D reconstruction~\eg\cite{TreePartNet21}, organ segmentation~\eg\cite{mirande2022graph}, or skeletonization~\eg\cite{dobbs2023smart}. 

We propose a method to infer a parameterized representation given as input a 3D scan of a plant. This is achieved by learning a shape space of 3D plants that captures both the plant's structure and the parametric 3D shape of all plant organs. The resulting representation can be directly used to reconstruct the 3D geometry of the plant with labels. This offers the key advantage of allowing to solve a variety of tasks, while allowing for complex plant architectures. 

Learning shape spaces of plants is under-explored due to challenges arising from variation in both plant structure and organ shape. A recent work proposes a first solution~\cite{cheng2025demeter}, while relying on large captured annotated datasets for training and requiring pre-segmented input for inference, both of which are costly and error-prone. In contrast, we propose a method trained purely on synthetic data and show its applicability to raw 3D scan acquisitions without annotations. 

To learn a shape space of 3D plants, we leverage biologically inspired procedural models made of recursive rules that describe plant development. Procedural models inform both the design of our neural network that learns a shape space of 3D plants and the generation of large amounts of synthetic training data. We use Lindenmayer-systems (L-systems) to represent plants in a binary axial tree form~\cite{Prusinkiewicz1990TheAB}. In L-systems, each plant architecture is represented as a bracketed string of parameterized modules, the L-String. We use a recursive neural network to model the recursive nature of the L-string. Without loss of generality, we restrict plant representations to binary trees, which allows the use of recursive auto-encoders trained on (binary) L-Strings. In this way, we learn both branching structure and shape distribution of plant organs from synthetic data. 
To generalize to acquired 3D scans of plants, we learn a mapping from simulated 3D scan data to the learned shape space. For robustness, we simulate acquisition noise on the virtual plants. 

We evaluate our method on \textit{Chenopodium album} plants, an annual plant, in early growth stages. Chenopodium is an ideal plant for our tests as it is a small plant (as opposed to trees) while still displaying a complex branching architecture and has small leaves that partially occlude each other, but not excessively. In addition, a collection of real scans are readily available at different stages of development~\cite{mirande_2022_6962994}. We thus created for this plant a large dataset of 3D virtual plants represented as L-Strings, with associated 3D point clouds simulated with different types of acquisition noise.

We use our method to perform 3D reconstruction, skeleton extraction and plant organ segmentation. Our experiments show results on-par with strong baselines, and robust to noisy and partial inputs.   

The main contributions of this work are:
\begin{itemize}
    \item An approach to infer a parameterized representation of a plant from a 3D scan based on a shape space of 3D plants learned using a recursive neural network. 
    \item 3D plant reconstruction, skeletonization and segmentation, with performance on-par with strong baselines.
    \item A dataset of virtual plants, in the form of L-Strings, that represents instances of Chenopodium Album plants in early growth stages, at~\href{https://doi.org/10.57745/7STDEK}{doi.org/10.57745/7STDEK}.
    \item A demonstration that learning based exclusively on virtual plants can be generalized to fit real plants.
\end{itemize}

\section{Related Work}
\label{sec:realated_work}

Existing works to infer high-level information of plants belong to two main categories: inverse plant modeling and task-specific methods. Our work is between these categories, since it infers a parameterized plant representation, which includes the branching structure and 3D geometry of the plant. This infers a step of a procedural model and can be seen itself as a step towards inverse plant modeling.

A notable exception to our categorization is Demeter~\cite{cheng2025demeter}, a work close in spirit to ours. Demeter learns a parametric plant model and demonstrates its use for reconstruction tasks. Unlike our method, Demeter requires manually annotated 3D scans for training and uses an elaborate three-step method for reconstruction, which requires pre-segmented input. In contrast, our method is trained on virtual data, which is cheaper to obtain, and directly infers a parametric representation without costly data pre-processing.

\subsection{Inverse Plant Modeling}

Procedural models generate plants following growth rules. In contrast, inverse plant modeling aims to find the rules that allow to generate a given plant. Guo~\etal~\cite{guo2020inverse} and {\v{S}}t{\'{a}}va~\etal~\cite{vst2010inverse} infer procedural modeling rules generated using an L-system from 2D images of leafless branching structures. {\v{S}}t{\'{a}}va~\etal~\cite{stava2014inverse} introduce a parametric procedural model in 3D specific to trees. None of these methods is applicable to annual plants with leaves. 
More recently, Lee~\etal~\cite{10.1145/3627101} used transformers to learn a generative model to represent L-system rules, without the aim of reconstructing a precise instance of a given real plant input. CropCraft~\cite{zhai2024cropcraftinverseproceduralmodeling} performs inverse procedural modeling taking as input a collection of 2D images, and not 3D scans.
\subsection{Task Specific Methods}

\begin{table}[h]
    \centering
    {\scriptsize
    \begin{tabular}{|c||c|c|c|c|c|}
    
         \hline
         Work & Recon- & Skeletoni- & Segment- & Annual & Direct \\
         & struction & zation & ation & plants & infer-\\
         & & & & & ence\\
         \hline
         \hline
         {\tiny Gonzalez~\etal~\cite{gonzalez2023tree}} & \textcolor{green}{\cmark} & \xmark & \textcolor{green}{\cmark}  & \xmark & \xmark\\
         \hline
        {\tiny Liu~\etal~\cite{TreePartNet21}} & \textcolor{green}{\cmark} & \textcolor{green}{\cmark} & \textcolor{green}{\cmark}  & \xmark & \textcolor{green}{\cmark}\\
          \hline
          {\tiny Du~\etal~\cite{du2019adtree}}& \textcolor{green}{\cmark} & \textcolor{green}{\cmark} & \xmark & \xmark & \textcolor{green}{\cmark}\\
          \hline
          
          {\tiny Linvy~\etal~\cite{livny2010automatic}}& \textcolor{green}{\cmark} & \textcolor{green}{\cmark} & \xmark & \xmark & \textcolor{green}{\cmark}\\
          \hline
         {\tiny Preuksakarn~\etal~\cite{plant-recon}} & \textcolor{green}{\cmark} & \textcolor{green}{\cmark} & \xmark &  \xmark & \textcolor{green}{\cmark}\\
            \hline
         {\tiny Parsad~\etal~\cite{prasad2022deep}}& \textcolor{green}{\cmark} & \xmark & \xmark & \textcolor{green}{\cmark} & \textcolor{green}{\cmark}\\
          \hline
        {\tiny Dobbs~\etal~\cite{dobbs2023smart}}& \xmark & \textcolor{green}{\cmark} & \xmark & \xmark & \textcolor{green}{\cmark}\\
         \hline
         {\tiny Chaudhury~\etal~\cite{Chaudhury2020.02.15.950519}} & \xmark & \textcolor{green}{\cmark} & \xmark & \textcolor{green}{\cmark} & \xmark\\
           \hline
         {\tiny Yan~\etal~\cite{5246837}} & \textcolor{green}{\cmark} & \textcolor{green}{\cmark} & \textcolor{green}{\cmark} &
         \xmark & \textcolor{green}{\cmark}\\
          \hline
         {\tiny Meyers~\etal~\cite{Meyer_2023}} & \xmark & \textcolor{green}{\cmark} & \textcolor{green}{\cmark} & \xmark &  \textcolor{green}{\cmark} \\
          \hline
          {\tiny Mirande~\etal~\cite{mirande2022graph}} & \xmark & \xmark & \textcolor{green}{\cmark} & \textcolor{green}{\cmark} & \textcolor{green}{\cmark}  \\
          \hline
         {\tiny Turgut~\etal~\cite{turgut2022segmentation}}& \xmark & \xmark & \textcolor{green}{\cmark} & \textcolor{green}{\cmark} & \textcolor{green}{\cmark}\\
          \hline
          {\tiny Turgut~\etal~\cite{TURGUT2022138}}& \xmark & \xmark & \textcolor{green}{\cmark} & \textcolor{green}{\cmark} & \xmark\\
          \hline
          {\tiny Wahabzada~\etal~\cite{2015article}}& \xmark & \xmark & \textcolor{green}{\cmark} &  \textcolor{green}{\cmark} &\textcolor{green}{\cmark} \\
          \hline
          {\tiny Li~\etal~\cite{LI2022243}}  & \xmark & \xmark & \textcolor{green}{\cmark} & \textcolor{green}{\cmark } & \textcolor{green}{\cmark} \\
          \hline
          {\tiny Li~\etal~\cite{psegnet2022}}  & \xmark & \xmark & \textcolor{green}{\cmark} & \textcolor{green}{\cmark } &  \textcolor{green}{\cmark} \\
          \hline
          {\tiny Cheng~\etal~\cite{cheng2025demeter}} & \textcolor{green}{\cmark}    &\textcolor{green}{\cmark} & \textcolor{green}{\cmark} &  \textcolor{green}{\cmark} & \xmark\\
          \hline
          \hline
         Ours   & \textcolor{green}{\cmark}    &\textcolor{green}{\cmark} & \textcolor{green}{\cmark} &  \textcolor{green}{\cmark} &  \textcolor{green}{\cmark}\\
         \hline
    \end{tabular}%
    }    
    \caption{Positioning of our method w.r.t.~the ability to perform 3D reconstruction, skeletonization and segmentation, applicability on annual plants, and ability to perform inference directly without requiring pre-segmentation of the input data.}
    \label{tab:positioning}
\end{table}

Existing plant phenotyping methods from 3D point clouds mainly focus on three tasks: 3D reconstruction, extracting a skeleton of the branching structure, and segmentation.

\paragraph{3D Reconstruction}
While some works reconstruct 3D plants from a single 2D image~\cite{liu2025boxplant}, we focus on reconstructing plant geometry from an input 3D point cloud. 
This 3D reconstruction problem is challenging  as plants contain thin structures and self-occlusions.

In computer graphics, Gonzalez~\etal~\cite{gonzalez2023tree} reconstruct urban trees by computing a mesh representing the tree trunk, estimating the volume and density of the canopy, and filling the canopy with generated leaves. The branching structure of the plant is ignored. 
Another line of work focuses on branching structures of trees~\cite{du2019adtree, plant-recon, 5246837, TreePartNet21, 10.1145/1289603.1289610} by first extracting a skeleton and subsequently reconstructing a mesh per branch. Foliage is not reconstructed. 

Closest to our work, Prasad~\etal~\cite{prasad2022deep} compare different implicit reconstruction algorithms on 3D point clouds of a plant. We compare our method experimentally to SIREN~\cite{sitzmann2020implicitneuralrepresentationsperiodic}, the best performing method tested in this work.

\paragraph{Skeletonization}
One line of work, discussed above, extracts skeletons as one part of a pipeline to reconstruct the plant~\cite{du2019adtree, plant-recon, 5246837, TreePartNet21, 10.1145/1289603.1289610}. 
Other approaches use optimization-based methods. Meyer~\etal~\cite{Meyer_2023} use a point contraction algorithm to extract skeletons of leafless trees. Graph-based approaches have shown to extract skeletal structures even in the presence of noise~\cite{10.1145/1289603.1289610, livny2010automatic}. A recent deep learning-based approach estimates the medial axis of a tree~\cite{dobbs2023smart}.
Given a point cloud of a plant with an initial extracted skeleton, Chaudhury~\etal~\cite{Chaudhury2020.02.15.950519} refine the skeleton.

All discussed methods require leafless input. When leaves are present, skeletal structures are estimated inside the leaves, resulting in noisy output. Our method allows to extract skeletons with or without leaves. We experimentally compare our method to one successful approach from each category for which code is available~\cite{10.1145/1289603.1289610,livny2010automatic, Chaudhury2020.02.15.950519}.

\paragraph{Segmentation}
Both learning and optimization-based techniques can segment a plant into organs. Mirande~\etal~\cite{mirande2022graph} propose a graph-based optimization approach with botanical knowledge refinement for semantic and instance segmentation. Wahabzada~\etal~\cite{2015article} present an unsupervised data-driven method. For supervised learning methods, a benchmark for plant organ segmentation~\cite{turgut2022segmentation} has been released based on the ROSE-X dataset~\cite{Dutagaci2020ROSEXAA}. More complex deep learning architectures designed for organ segmentation lead to accurate results~\cite{LI2022243, TURGUT2022138, psegnet2022}. We compare our method to the two generally applicable approaches~\cite{LI2022243, psegnet2022}.

\paragraph{Positioning}
Table~\ref{tab:positioning} shows that only few works perform 3D reconstruction, skeletonization and segmentation~\cite{TreePartNet21, 5246837,cheng2025demeter}. Two of these~\cite{TreePartNet21, 5246837} are limited to big trees with trunks and foliage, and not applicable on small annual plants. The third method~\cite{cheng2025demeter} takes 2D images as input and performs segmentation as pre-processing, and is therefore not comparable to our method. 

\section{Background}

This section provides background on plant modeling and recursive neural networks.

\subsection{Plant Modeling}\label{plant-model}

Plant modeling aims to find a mathematical model that describes the complex geometry and growth rules for a species of plants. Moreover, plant models respect biological rules to allow realistic simulation of the plant's appearance and behavior in different environmental conditions~\cite{Godin.2005}.

Procedural methods are among the most commonly used plant modeling approaches, including L-systems~\cite{Prusinkiewicz1990TheAB}, the space colonization algorithm~\cite{space}, and included in interactive tools such as SpeedTree~\cite{speedtree}. In our work, we use procedural plant models to generate training and test data. 

Our implementation uses an L-system-based procedural approach~\cite{Prusinkiewicz1990TheAB} to generate plant data, a classical rule based plant modeling technique, that is particularly convenient due to the availability of tools such as L-Py~\cite{boudon2012py}. In L-systems, a plant is represented as a string of symbols, called L-String, possibly bearing geometrical or biological parameters. Similarly to natural or computer language grammars, an L-system consists of a set of rewriting rules that define how plant components represented by L-String symbols change as time proceeds, by specifying how symbols get replaced by combinations of other symbols~\cite{Prusinkiewicz1990TheAB, Godin.2005}. At each step, the rules replace the symbols in parallel, resulting in a new L-String representing the next plant state.

L-Strings of plants mainly consist of two types of symbols: \emph{modules} and \emph{brackets}. Modules can refer to various plant parts \eg stems, leaves, flowers, etc. Opening and closing brackets indicate the start and the end of every branch in the plant.
Two successive modules in an L-string have a parent-child relationship. Open brackets allow for a module to have more than one child, and for different children to have siblings relationships. However, each parent has at most one special child that corresponds to its successor on the same plant axis. Such L-Strings encode axial trees~\cite{Prusinkiewicz1990TheAB} that represent the plant's architecture. 
Modules in L-Strings can have parameters that give information about a plant organ, which allows an L-String to encode a plant's geometry in addition to its topology, defined as its structure. 
\subsection{Recursive Neural Networks}

Recursive Neural Networks (RvNN)~\cite{costa2003towards} are deep neural networks that apply the same network architecture recursively on structured input. The term \emph{recursive} refers to the network being applied to the output of step $i-1$ during step $i$. The recursivity of RvNNs allows for input of varying size, as the network is applied bottom-up from leaves to root.

RvNNs have been applied to different domains~\cite{socher2010learning,socher2011parsing} including natural language processing~\cite{chuan-an-etal-2018-unified}, 3D generation~\cite{li2019grains, 10018465,10.1145/3072959.3073637}, blood vessel synthesis~\cite{10.1007/978-3-031-43907-0_7}, 3D shape structure recovery~\cite{Niu_2018_CVPR}, and segmentation~\cite{Yu_2019_CVPR}.

Of particular interest for our work are recursive auto-encoders for binary trees~\cite{10.1145/3072959.3073637}. A recursive auto-encoder learns on a binary tree structure, where all nodes can be represented in a latent space of dimension $\dimension_{\latentspace}$. The encoder follows the tree structure and recursively merges pairs of inputs to form a new point in the same latent space until the full tree is represented as a single latent point. Inversely, the decoder recursively decodes a single point into two points in the same latent space until the full tree structure is decoded.

\section{Dataset}
\label{sec:data}

To train and evaluate our method, we design and generate a synthetic dataset of corresponding L-String and point cloud pairs of the Chenopodium Album plant. First, we generate the L-Strings using L-Py platform~\cite{boudon2012py}, with L-system production and geometric interpretation~\cite{prusinkiewicz1986graphical}, that we optimize to generate realistic Chenopodium virtual plants. We defined different time functions for the different plant parameters that guide the plant growth for a range of $[8,14]$ days to get Chenopodium plants in early stages. Then, we used the labeled points sampling method from~\cite{chaudhury20203d} to obtain the corresponding point clouds. 

The L-Strings of the generated Chenopodium Album plants consist of 5 different modules: stem, cotyledon, petiole, leaf and branch. For each module, there is a different set of parameters that describe the organ represented by the module. The modules' parameters are as follows:
\begin{itemize}
    \item Stem: diameter, length, growing angle, bending angle and phyllotaxis angle.
    \item Cotyledon: angle, length, nerve curvature factor.
    \item Petiole: starting diameter, ending diameter, angle, length, and elasticity factor.
    \item Leaf: nerve curvature factor, length and width.
    \item Branch: branching angle and elasticity factor.
\end{itemize}
The dataset contains plants of different shapes and structures. We balance the different structures in the dataset by fixing the number of different plants for each structure. We generated plants of 10 different structures with 100 different plants per structure, resulting in a dataset of 1000 pairs of L-Strings and point clouds.

\section{Method}

Our goal is to infer an L-String $\lstring$ from an input 3D scan of a plant represented as point cloud $\pointcloud$.
Learning a direct regression from the space of 3D point clouds to the space $\lstringspace$ of L-Strings is difficult, as point clouds have varying numbers of points and are unstructured, and as L-Strings vary both in discrete (\ie number and type of modules) and continuous (\ie values of the angle and length parameters) ways.

\begin{figure}
    \centering
    \includegraphics[width=\linewidth]{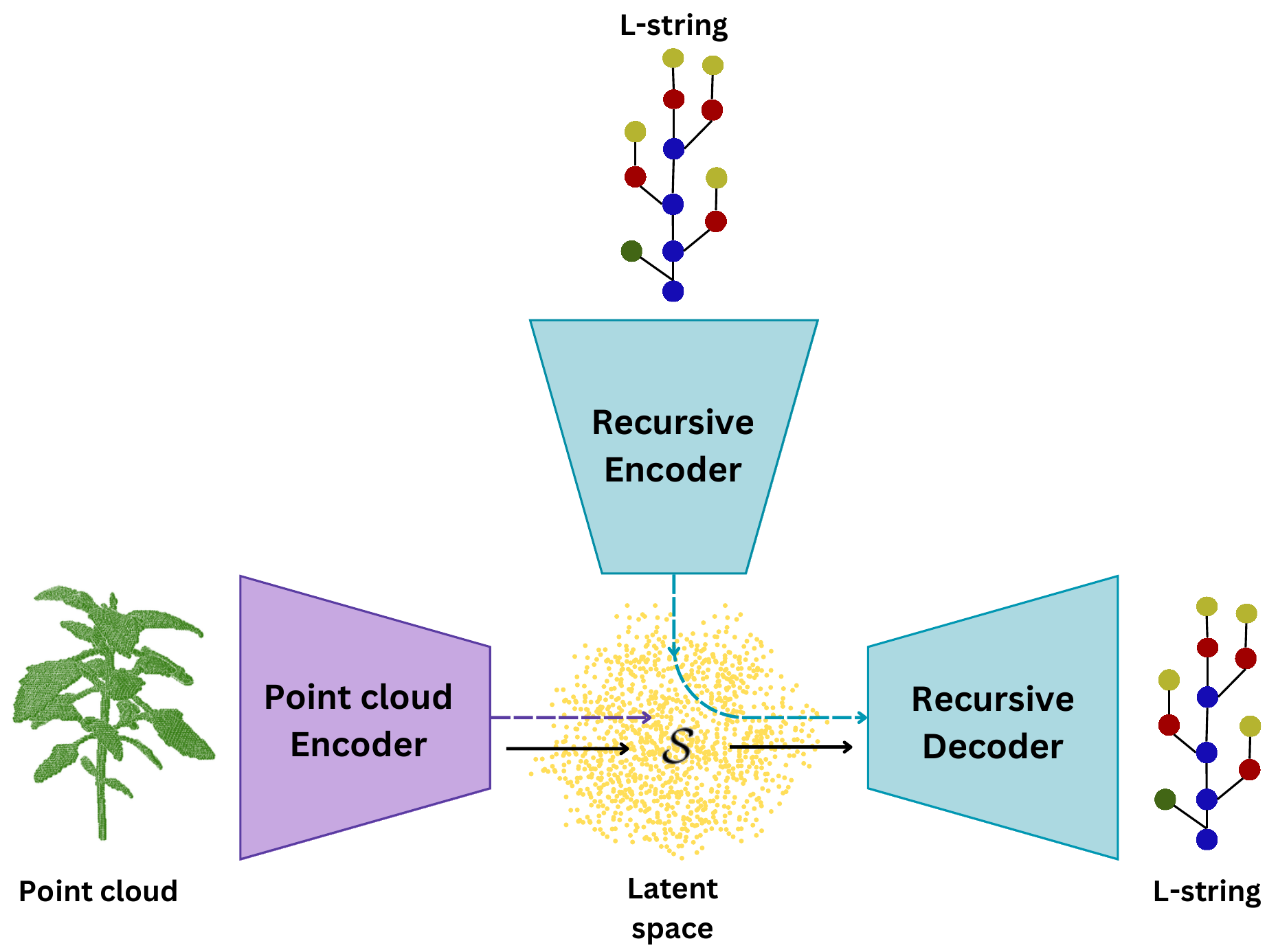}
    \caption{Overview: our method learns a latent space $\latentspace$, that allows the mapping of 3D point clouds to L-Strings. At inference, the point cloud is mapped to $\latentspace$ using the point cloud encoder on the left, and the resulting latent point allows to reconstruct the corresponding L-String using the L-String decoder on the right.}
    \label{fig:pipeline}
\end{figure}

To address this problem, we combine an RvNN learned on $\lstringspace$ with an encoder that maps a point cloud to a latent space $\latentspace$, as shown in Fig.~\ref{fig:pipeline}. During training, we first learn the latent space $\latentspace$ that allows to represent L-Strings of a fixed plant species, see Sec.~\ref{method:lstring}. Second, we train a neural network called point cloud encoder that maps an input point cloud $\pointcloud$ to a point $\latentcode \in \latentspace$, see Sec.~\ref{method:3d}. During inference, the input point cloud $\pointcloud$ is encoded in $\latentcode \in \latentspace$ using the point cloud encoder, and subsequently decoded to an L-String using the recursive decoder, see Sec.~\ref{method:infer}. 

\subsection{Representing L-Strings in Latent Space}
\label{method:lstring}

We aim to learn a latent space $\latentspace$ that represents instances of a plant species. The branching structure of the L-Strings may not always correspond to a binary tree structure. We therefore first simplify the L-Strings by summarizing the information of modules that always occur together in the plant species. The definition of co-occurring modules needs to be manually done once per species. The resulting combined modules are called nodes in the following, and we design the combination rules to guarantee a binary tree structure after combination. To simplify notations, we call the L-String with binary tree graph structure $\lstring \in \lstringspace$ in the following.

Our goal is to learn an encoder function $\encoderfunction : \lstringspace \to \latentspace$, and a decoder function $\decoderfunction : \latentspace \to \lstringspace$, such that $\lstring \approx \decoderfunction(\encoderfunction(\lstring)), \;\; \forall \lstring \in \lstringspace$.
Learning to encode different shape and structural information in latent points of fixed dimension is challenging. Inspired by~\cite{10.1145/3072959.3073637}, we use an RvNN to learn the hierarchical relations between the modules of the L-String, leveraging the recursive nature of plant structures~\cite{Godin.20103zn}.

To achieve this, each node of the tree graph is represented individually as point in latent space, $\latentcode_i \in \latentspace$, using a node auto-encoder. In a second step, each subtree of $\lstring$ is represented in $\latentspace$ using a recursive auto-encoder. This procedure is carried out recursively from the leaves to the root resulting in a latent point that represents the full $\lstring$ in $\latentspace$.

\subsubsection{Node Auto-encoders} 

Each type of node has a different set of parameters,~\eg angles, widths, radii. This results in nodes that have different dimensionality in general, and that are not directly comparable. To allow nodes with different numbers of parameters to be used as input in an RvNN, we first map the information of each node to $\latentspace$. To achieve this, a node encoder-decoder pair is learned for each node type. In the following, let $\nodeencoderi$ and $\nodedecoderi$ denote the encoder and decoder for each type of node, with $i=1,\ldots,\numberNodeType$ and $\numberNodeType$ the number of types of nodes.
Both $\nodeencoderi$ and $\nodedecoderi$ consist of one fully connected linear layer, followed by a $tanh$ activation.
\subsubsection{Recursive Auto-encoders}

\begin{figure}
    \centering
    \includegraphics[width=\linewidth]{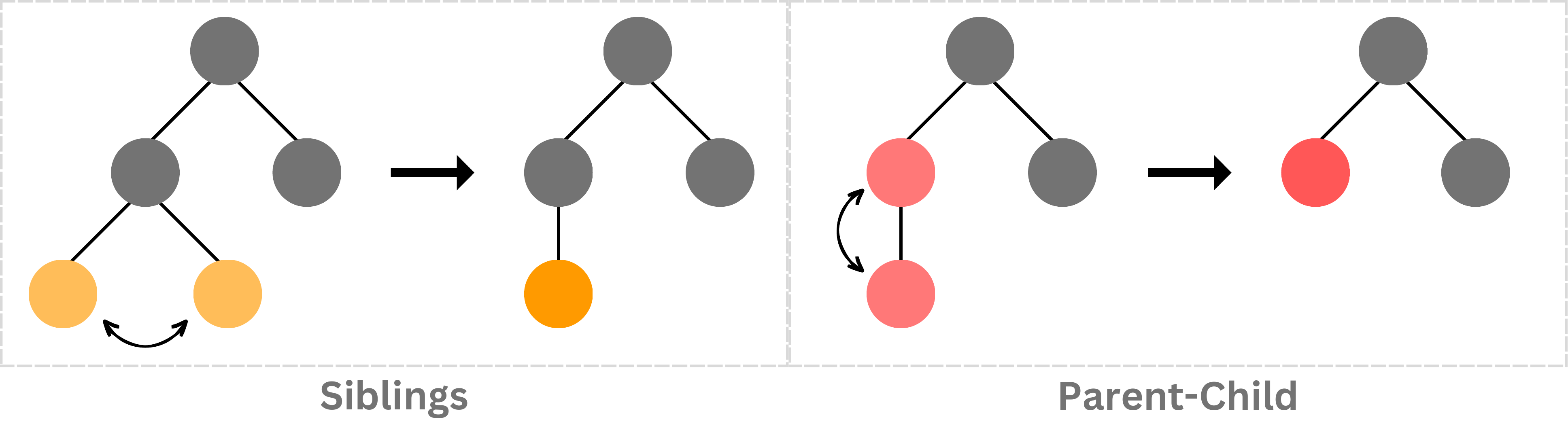}
    \caption{The two criteria to merge/split points in latent space shown on an example tree. Merging is applied recursively in a bottom-up manner until the whole tree is merged into one point, while the splitting performs the inverse operation.}
    \label{fig:criteria}
\end{figure}  

\begin{figure}
    \centering
    \includegraphics[width=\linewidth]{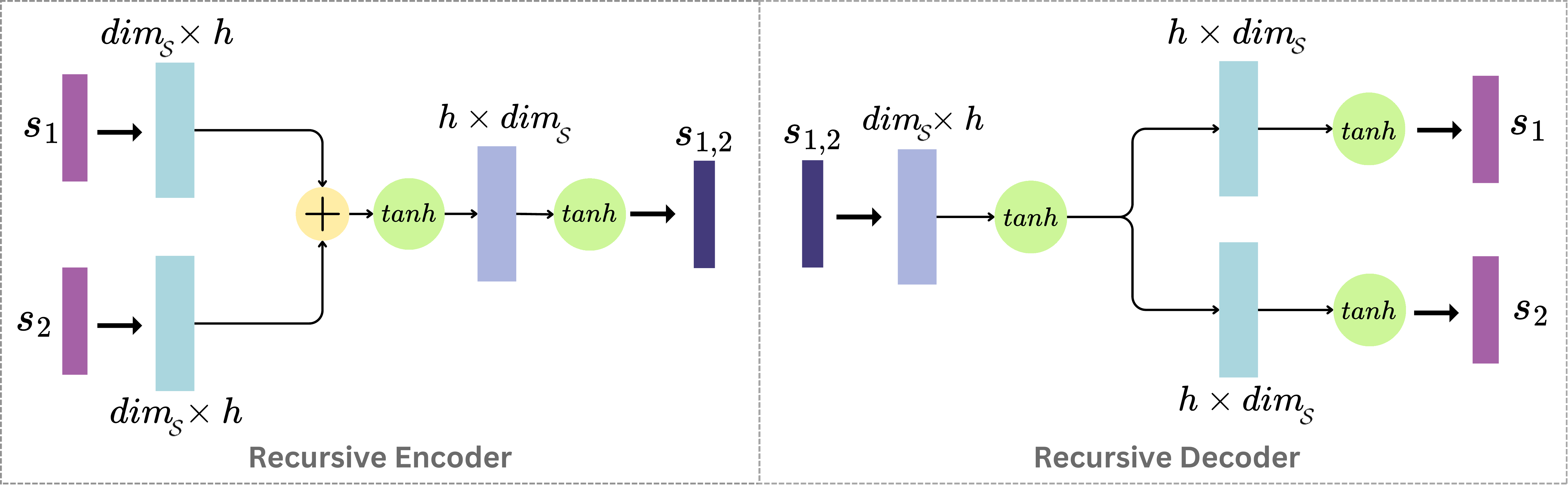}
    \caption{The network architectures used for the recursive encoder and decoder. A binary tree structure is recursively encoded into latent space $\latentspace$. Latent points $\latentcode_1 \in \dimension_{\latentspace}$ and $\latentcode_2 \in \dimension_{\latentspace}$ are merged into $\latentcode_{1,2} \in \dimension_{\latentspace}$ by the encoder. Symmetrically, the decoder splits $\latentcode_{1,2}$ into two latent points $\latentcode_1$ (the special child in the axis) and $\latentcode_2$. $\dimension_{\hidden}$ denotes the dimension of the hidden layer.}
    \label{fig:recursive_arch}
\end{figure}

After applying node encoding, an L-String is represented as a binary tree where all nodes are individually represented as points in $\latentspace$. This is the input to an RvNN auto-encoder. 
To recursively merge or split nodes, we design encoder-decoder pairs based on the relationship of the input nodes in the tree. Two nodes to be merged can either be siblings or have a parent-child relationship in the tree. 
Based on this observation, we consider the two merging/splitting criteria shown in Fig.~\ref{fig:criteria}. For each criterion, we learn one recursive encoder-decoder pair. Both the sibling encoder-decoder $(\siblingencoder, \siblingdecoder)$ and the parent-child encoder-decoder $(\pcencoder,\pcdecoder)$ are implemented as shown in Fig~\ref{fig:recursive_arch}.

\subsubsection{Auxiliary Classifiers}

In addition to the encoder-decoder pairs, it must be decided how to properly split the points to reconstruct the L-String structure. Each point in $\latentspace$ is either a result of a merging process by one of the recursive encoders ($\siblingencoder$, $\pcencoder$), or it represents an individual node. For the RvNN to decode structural information, it needs to learn to choose the appropriate decoder ($\siblingdecoder$, $\pcdecoder$),  or to end the recursive splitting. To predict the right decoder, we associate classes for the different splitting options (siblings, parent-child, stop) and jointly train a classifier $\classifierSplit$ on them.

After splitting, each latent point needs to be decoded with an appropriate decoder $\nodedecoderi$ to reconstruct a node of the L-String. For that, a second classifier is trained to predict the node type for a latent point $\latentcode$. This classifier is called $\classifierNode$. 
Both classifiers consist of a multi-layer perceptron with a single hidden layer and a $tanh$ activation. 

\subsubsection{Training}

All auto-encoders are trained by minimizing a reconstruction loss at the node level. For node $\node$ in L-String $\lstring$, with a set of parameters, the reconstruction loss $\reconstructionloss(\node)$ is the mean squared error between the input node parameter vector that contains the plant part information and its reconstruction. The parameters' weights are different in the loss.
The reconstruction loss for the full L-String is 
\begin{equation}
    \reconstructionloss = \sum_n\reconstructionloss(n).
\end{equation}

The classifiers are trained with softmax classification and cross entropy loss. $\classifierSplit$ takes a latent point $\latentcode$ as input, and predicts one of the three classes: parent-child split, siblings split or leaf node (no split). $\classifierNode$ predicts one of $\numberNodeType$ possible node types. We denote the cross entropy losses used to train these classifiers by $\splitloss$ and $\nodeloss$, respectively.

Finally, the total loss that is used for training the recursive L-String auto-encoder is
\begin{equation}
\label{eq:loss}
    \totalloss = \reconstructionloss
    +\splitloss + \nodeloss.
\end{equation}

\subsection{Point Cloud Encoder}
\label{method:3d}

After learning latent shape space $\latentspace$, 3D plants can be represented as $\latentcode \in \latentspace$, and $\latentcode$ can be decoded into a parametric L-String representation $\lstring$. To infer an L-String $\lstring$ from an input point cloud $\pointcloud$, we learn a mapping function $\encoderfunctionpoints$ from the space of point clouds to $\latentspace$ using a PointNet~\cite{DBLP:journals/corr/QiSMG16}.

To train $\encoderfunctionpoints$, we take advantage of paired input data, containing both the point cloud $\pointcloud$ and its corresponding L-String $\lstring$. Passing $\lstring$ through the recursive L-String encoder produces a latent point $\latentcode$. By applying $\encoderfunctionpoints$ on $\pointcloud$, we obtain $\hat{\latentcode}$. The training optimizes the loss
\begin{equation}
    \pointcloudloss = \sum_j \left(\hat{\latentcode}_j - \latentcode_j \right)^2,
\end{equation}
where $j$ loops over all training samples. This loss encourages the point cloud encoder $\encoderfunctionpoints$ to represent $\pointcloud$ at the location $\hat{\latentcode} \in \latentspace$ that represents its corresponding L-String. 

\subsection{Inferring L-Strings from Input Point Clouds}
\label{method:infer}
At inference, the input is an unstructured point cloud $\pointcloud$. This input is encoded in $\latentspace$ using the point cloud encoder, and the resulting point is decoded using the recursive decoder as $\lstring = \decoderfunction(\encoderfunctionpoints(\pointcloud))$.

Errors in the predicted module parameters of the reconstructed L-String can lead to cumulated errors on the plant reconstruction. For example, errors in predicting the angle of a stem that is located in the bottom of the plant can lead to deviation in the plant growth axis along the main stem. To avoid such deviations, we align the reconstructed plant with the input in a test-time optimization framework. We optimize on parameters of the main stem modules starting from the ones at the bottom of the plant and going up. This optimization is done on 3D angle and length parameters. We then optimize on the parameters of the petiole modules that define the length and elasticity, and then  on the parameters of leaf modules that define the leaf size and curvature. All the modules are optimized \wrt the bidirectional Chamfer Distance between the reconstructed plant and the input point cloud, and two optimization iterations from bottom to top are performed.
This results in a parametric L-String representation $\lstring$ that allows for various downstream phenotyping tasks. In this paper, we focus on the following tasks.

\textbf{3D Reconstruction} can be solved by applying the geometric interpretation rules on $\lstring$ to retrieve the 3D plant. 

\textbf{Skeletonization} is solved by applying geometric interpretation rules on $\lstring$ to reconstruct all stems and optionally main veins of leaves with minimal width.

\textbf{Segmentation} is solved by applying geometric interpretation rules on $\lstring$ and keeping the labels of the organ types. The labels are propagated to $\pointcloud$ by assigning each point in $\pointcloud$ the most frequently assigned label among its $k$ nearest neighbors in the annotated point cloud corresponding to $\lstring$.

\section{Evaluation}

In this section we evaluate our method for the three common phenotyping applications 3D reconstruction, 3D skeleton extraction, and segmentation. For each application, we outline an evaluation protocol and compare to strong baselines. All methods are run on a Quadro RTX 5000 GPU. Implementation details are in the supplementary materials.

The dataset in Sec.~\ref{sec:data} is split into training ($80\%$), validation ($10\%$) and test ($10\%$) sets. This dataset of realistic Chenopodium Album plants with ground truth parametric representation can serve as benchmark for plant reconstruction, segmentation and skeletonization tasks. 

We evaluate on the test set with clean point clouds, point clouds with simulated Gaussian noise, and monocular depth images of the set, to show our model's robustness. 
To test generalization to real data, we test our method on 5 scans of real Chenopodium Album plants from Mirande~\etal~\cite{mirande_2022_6962994}.

\begin{table*}[ht]
{\footnotesize
\centering
    \begin{tabular}{lcccccccc}
    \toprule
    \textbf{} & \textbf{Accuracy} & \textbf{Completeness}  & \textbf{Size} & \textbf{Time} & \textbf{ \# Comp.} & \textbf{LN Accuracy} & \textbf{LAI Accuracy} & \textbf{Topology}\\
    \midrule
    \textbf{Clean}& & & & & & & & \\
    SIREN& $\mathbf{0.0012}$ & $\mathbf{0.0006}$&$780.88$ KB & $7\,min\,14\,s$ & $33$ & \xmark & \xmark & \xmark\\
    Ours& $0.0059$ & $0.0090$ & $\mathbf{17.56}$ \textbf{KB} & $\mathbf{4\,min\,1\,s}$ & $\mathbf{1}$ &$\mathbf{98\%}$ & $\mathbf{93\%}$ & $\mathbf{75\%}$  \\
    \midrule
    \textbf{Noisy}&  & & & & & & & \\
    SIREN& $0.0121$&$\mathbf{0.0008}$  & $780.88$ KB&$7\,min\,27\,s$ & $268$ & \xmark & \xmark & \xmark\\ 
    Ours & $\mathbf{0.0054}$ & $0.0074$ & $\mathbf{17.56}$ \textbf{KB}& $\mathbf{3\,min\,56\,s}$ & $\mathbf{1}$& $\mathbf{98\%}$& $\mathbf{93\%}$&$\mathbf{78\%}$ \\
    \midrule
    \textbf{Depth maps}& & & & & & & & \\
    SIREN&$\mathbf{0.0013}$ & $\mathbf{0.0022}$&$780.88$ KB &$6\,min\,44\,s$& $49$ &\xmark &\xmark &\xmark \\
    Ours& $0.0060$& $0.0089$& $\mathbf{17.57}$ \textbf{KB}& $\mathbf{3\,min\,47\,s}$ & $\mathbf{1}$ & $\mathbf{98\%}$ &$\mathbf{94\%}$ &$\mathbf{78\%}$ \\
    \bottomrule
    \end{tabular}
    \caption{Comparison to SIREN \cite{sitzmann2020implicitneuralrepresentationsperiodic} for 3D reconstruction on clean and noisy point clouds and depth maps. We compare geometric measures (accuracy, completeness), the model's size and inference time, number of connected components of output (\# Comp.), and phenotyping measures leaf number (LN) accuracy, leaf area index (LAI) accuracy, and topology. \xmark: the method cannot output the information.}
    \label{tab:reconstruction}

}
\end{table*}

\subsection{3D Reconstruction}

We quantitatively evaluate our method using eight complementary evaluation measures. The first two are \textit{accuracy} and \textit{completeness}, which are commonly used metrics assessing geometric alignment. Accuracy is the unidirectional Chamfer distance from the reconstructed surface to the ground truth, and completeness is the unidirectional Chamfer distance from the ground truth to the reconstructed surface. The next two measures are the \textit{output representation's size} and the \textit{inference's running time}, which measure the efficiency and compactness of the methods. We further report the \textit{number of connected components} of the output meshes. We also assess the ability of models to predict correctly the \textit{number of leaves}, defined as the mean percentage of matching leaf count between the reconstruction and the ground truth, and the mean percentage accuracy of \textit{leaf area index} (LAI), defined as the one-sided leaf area per unit ground surface area~\cite{watson_lai}. These two measures have been selected as they are commonly used in agronomy to calibrate crop models. Finally, \textit{topology accuracy} is the percentage of reconstructed plants whose topology is correct (measured using a tree-edit distance~\cite{ferraro-hal-00827474}).

\begin{figure}[t]
    \centering
    \includegraphics[width=\linewidth]{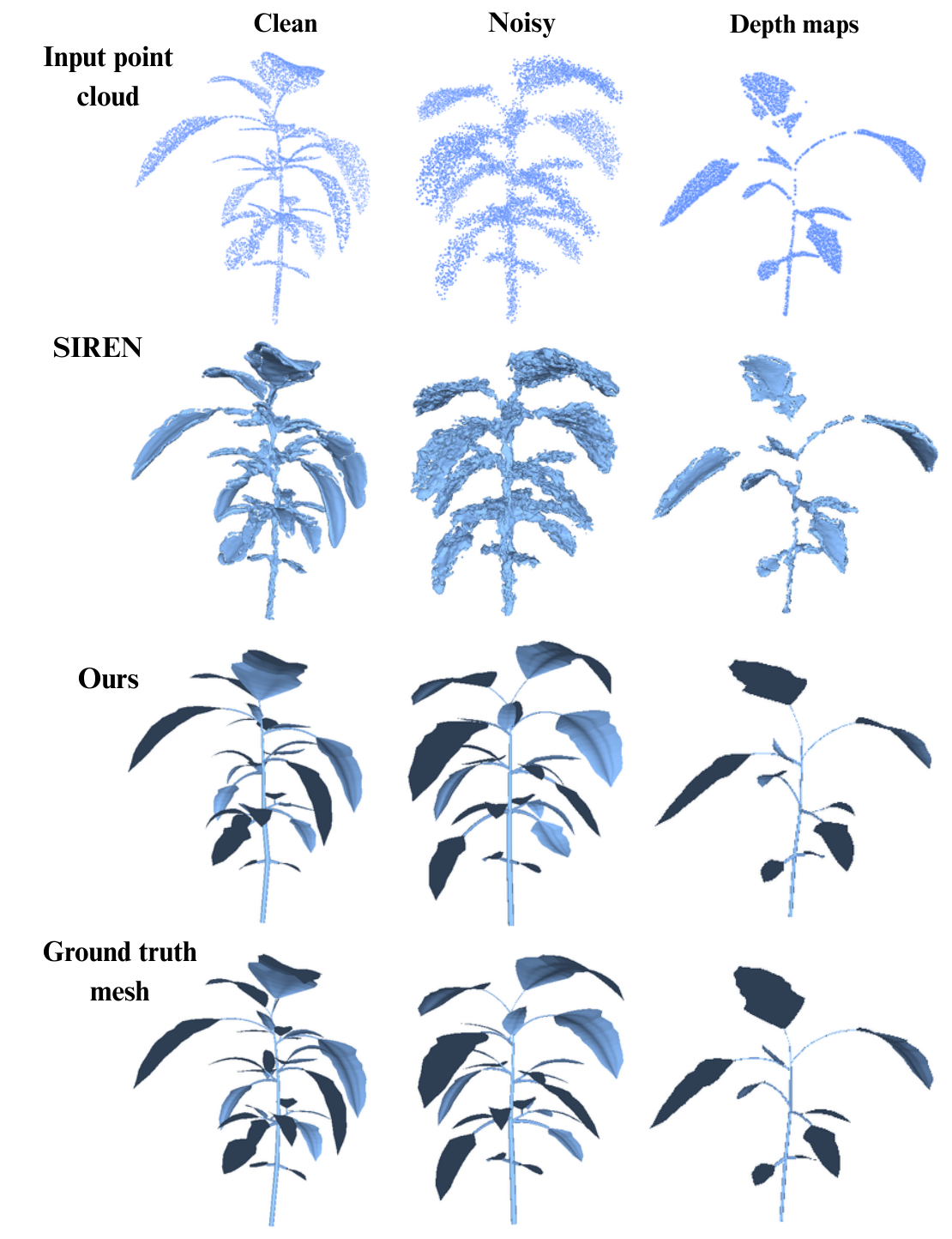}
    \caption{Comparison to SIREN~\cite{sitzmann2020implicitneuralrepresentationsperiodic} for 3D reconstruction.}
    \label{fig:reconstruction}
\end{figure}

We compare our method to SIREN~\cite{sitzmann2020implicitneuralrepresentationsperiodic}, an implicit neural method that uses sine activation functions to model continuous signals, including signed distance functions (SDFs). For 3D reconstruction, the surface is extracted as the zero level set of the SDF using Marching Cubes. SIREN was identified as the best-performing method to reconstruct plant geometry from point clouds by Prasad~\etal~\cite{prasad2022deep}.

Table~\ref{tab:reconstruction} shows the results. While SIREN performs well in terms of the 3D distance-based error measures accuracy and completeness for clean data, SIREN's performance degrades significantly for noisy data. Our method is more robust \wrt noise and missing data, and outperforms SIREN in case of noise. Our representation is one to two orders of magnitude more compact than SIREN's, and inference is twice as efficient. Unlike SIREN, which reconstructs a mesh without annotations, our method further allows to measure leaf number accuracy, LAI accuracy and topology, and achieves very high accuracy on all these measures.

\begin{figure}[t]
    \centering
    \includegraphics[width=\linewidth]{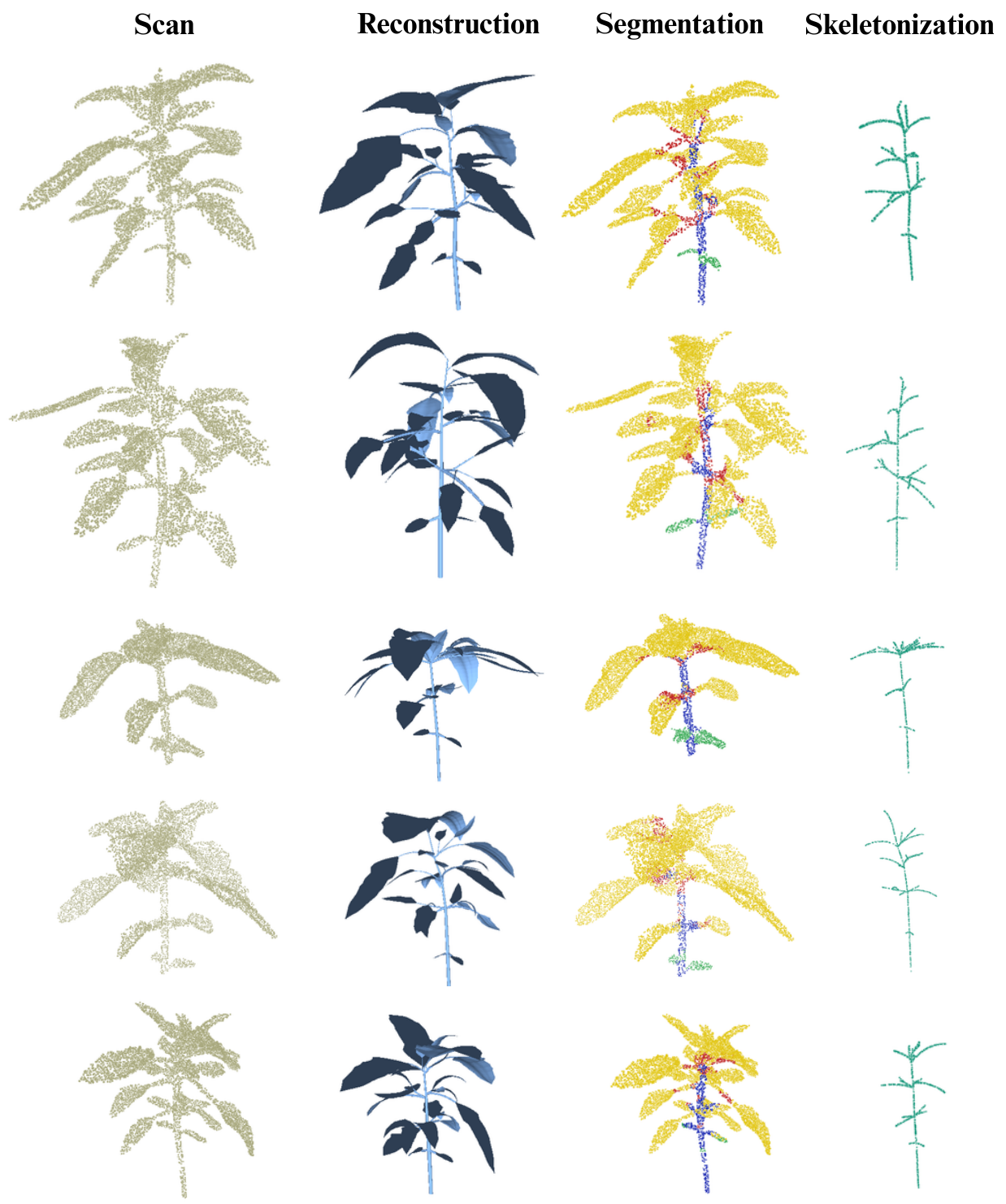}
    \caption{Our method's results on scans of real plants~\cite{mirande_2022_6962994}. Scans are reconstructed and segmented well overall, and the extracted skeletons closely follow the plants' topologies.}
    \label{fig:application_scan_data}
\end{figure}

Fig.~\ref{fig:reconstruction} shows 3D reconstructions for the three types of tests. Our reconstructions are smooth, similar to the ground truth, and robust \wrt input noise and missing data. In contrast, due to its computational strategy, SIREN reconstructs fragmented shapes close to the input point clouds that are globally dissimilar from the ground truth, especially for noisy input. Finally, Fig.~\ref{fig:application_scan_data} (second column) shows 3D reconstructions obtained by our method for scans of real plants. Despite local discrepancies, all plants are well reconstructed overall, which demonstrates that the method generalizes well from virtual to real plant architecture data.

\subsection{Skeletonization}
    
\begin{table}[ht]
    \centering
    {\footnotesize
    \begin{tabular}{l | c c c | c}
    \hline
     & \multicolumn{3}{c|}{\textbf{Full}} & \multicolumn{1}{c}{\textbf{Branch}}\\
         Method & Clean & Noisy & Depth images & Clean\\
        \hline
        Xu \etal & $0.0102$ & $0.0110$ & $0.0145$ & \xmark\\
        Chaudhury \etal & $0.0139$& $0.0154$& $0.0449$ & \xmark \\
        Livny \etal & $0.0257$& $0.0282$ & - & \xmark \\
        Ours & $0.0178$ & $0.0174$&  $0.0199$ & $0.0161$\\
        \hline
    \end{tabular}}
    \caption{Comparison for skeletonization \wrt Chamfer Distance. ``-'': method crashed due to numerical problems. \xmark: method's scale cannot operate on the branch level.}
    \label{tab:skeletonization}
\end{table}
We quantitatively evaluate 3D skeleton extraction using the commonly used bidirectional Chamfer Distance. 
We compare our method to two classical plant skeletonization baselines Livny~\etal\cite{livny2010automatic}, Xu~\etal~\cite{10.1145/1289603.1289610} and a stochastic skeleton refinement method Chaudhury~\etal~\cite{Chaudhury2020.02.15.950519} that takes a predicted skeleton as input and outputs a refined skeleton. The skeleton refinement is applied on the output skeletons of Xu~\etal~\cite{10.1145/1289603.1289610}. All these methods are designed to take scans of leafless plants as input, and their output cannot be adjusted to different scales. In contrast, our method can output skeletons with or without the main veins of the leaves.

\begin{figure}[t]
    \centering
    \includegraphics[width=\linewidth]{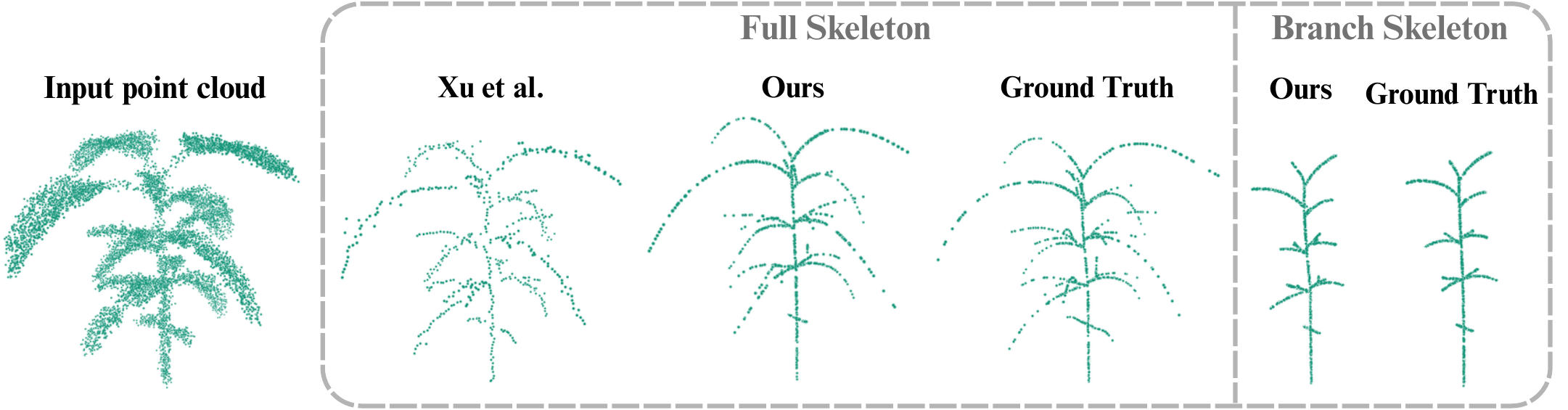}
    \caption{Comparison to Xu~\etal~\cite{10.1145/1289603.1289610} for skeletonization on noisy input. Ours is applicable at different scales, and can output a full or a branch skeleton, while achieving visually accurate results.}
    \label{fig:skeletons}
\end{figure}

Table~\ref{tab:skeletonization} shows the results for both the full skeleton that includes leaf veins and the branch skeleton that only includes the branching structure. Our method shows consistent performance on the different test sets, suggesting robustness to noise and missing parts in the input point clouds. 
Since baseline methods are not designed for input with leaves, they cannot output skeletons at different scales.
Fig.~\ref{fig:skeletons} shows a visual comparison on noisy input with the best method from Table~\ref{tab:skeletonization}, Xu~\etal~\cite{10.1145/1289603.1289610}, where our method achieves results close to the ground truth. More visual comparisons are in the supplementary materials. Fig.~\ref{fig:application_scan_data} (last column) shows skeletons extracted from scans of real plants. The skeletons closely follow the topology in the scans.

\subsection{Segmentation}

\begin{figure}[h]
    \centering
    \includegraphics[width=\linewidth]{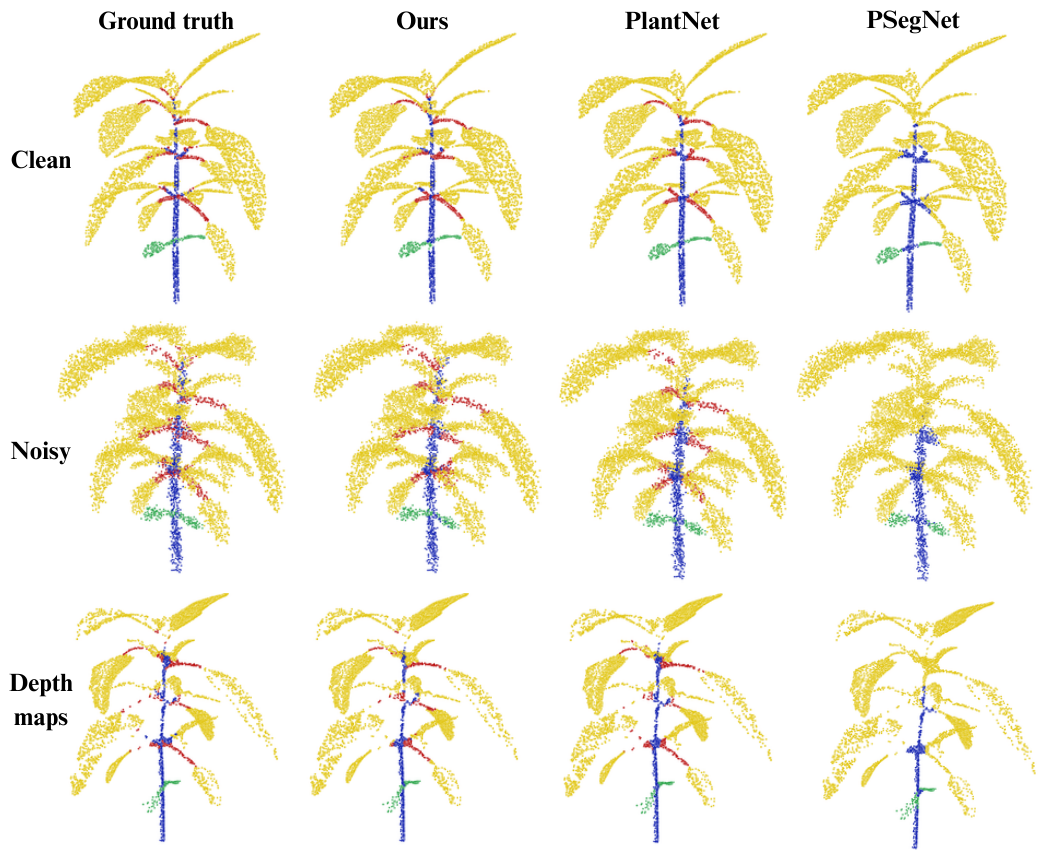}
    \caption{Comparison to PlantNet~\cite{LI2022243} and PSegNet~\cite{psegnet2022} for semantic segmentation on clean, noisy and partial point clouds.}
    \label{fig:semantic}
\end{figure}

We compare our method to the strong baselines PlantNet~\cite{LI2022243} and PSegNet~\cite{psegnet2022}. PSegNet provides labels on sparsely sampled point clouds, which we transfer to the full input point cloud using nearest neighbors similarly to our approach. Fig.~\ref{fig:semantic} shows the semantic segmentation results on examples from all test sets, where each type of plant organ is assigned a unique color. Our method is robust to noise and partial data, is on-par with the baselines, and outperforms PSegNet on the petioles segmentation. Quantitative comparisons on semantic segmentation and qualitative results on instance segmentation can be found in the supplementary materials.
Fig.~\ref{fig:application_scan_data} (third column) shows the segmentation of scans of real plants with our method. Note that all organs are well segmented overall.
\vspace{-10pt}
\section{Conclusion}

In this paper we have presented a data-driven method to infer parametric representations of plants \ie~L-Strings, from 3D unstructured point cloud input. This advancement was possible by leveraging procedural models both for designing the neural networks to learn a parametric shape space for 3D plants, and for generating synthetic training data. The L-String representation contains structural and geometric information that the input scans lack, and such information allows for multiple tasks like 3D reconstruction, skeletonization and segmentation at once, with performance on-par with strong baselines for each individual task. Our results have further shown that our method, trained purely on synthetic data, generalizes well to noisy scans of real plants. The main limitation is that a new model needs to be trained per species. This requires a L-System to generate training data, and a manual definition of nodes to obtain a binary tree structure. Furthermore, the method is limited to partially observable plant parts, preventing the modeling of mono-cotyledons e.g.~wheat and maize.

\vspace{-5pt}
\section{Acknowledgements}
We thank F.~H\'{e}troy-Wheeler and J.-S.~Franco for helpful discussions, and A.~Chaudhury and F.~Boudon for sharing their codes. This work was partially supported by French government funding managed by the National Research Agency under grant ANR-24-CE23-1586 (4DPlants).
\vspace{-15pt}
{
    \small
    \bibliographystyle{ieeenat_fullname}
    \bibliography{main}
}

\clearpage
\setcounter{page}{1}
\maketitlesupplementary



\section{Implementation Details}
We implemented our method using Pytorch on Quadro RTX 5000 GPU. We learn a latent space $\latentspace$ of dimension $\dimension_\latentspace = 64$ with the recursive encoder-decoder pair architecture in Fig.~4 in the main paper, of hidden layer size $\hidden_{recur} = 128$, and the classifiers with a hidden layer $\hidden_{class} = 128$. We optimize the weights of the network using Adam optimizer with a learning rate of 0.001, decaying by factor of 0.5 every 50 epochs, on batches of size 32.
The point cloud encoder is a PointNet point regression network that has the architecture shown in Fig.~2 in~\cite{DBLP:journals/corr/QiSMG16} without the optional feature transformation and with reduced dimensions for all layers. We train it with Adam optimiser on a learning rate of 0.001 on batches of size 32 without scheduling.
 The weights of the layers of all the trained networks are initialized using Xavier uniform initialization.

Our method works on binary trees, while the L-String trees in our dataset do not have a binary tree structure in general. We summarize the L-Strings into binary trees, by combining the modules that always occur together as individual nodes with concatenated parameter sets. In particular, we combine the two cotyledon modules into one \emph{Cotyledons} node with 6 parameters. Stems are always followed by a petiole and a leaf, therefore we combine them in one node called \emph{Stem} that has 13 parameters. In case of a branching, the branch module is combined with the stem, petiole and leaf modules to form a \emph{Branch} node of 15 parameters. Finally, the first stem module of the tree forms a node on its own called \emph{Root} of 5 parameters. Parameters that are drawn from a Gaussian distribution do not correlate with their parent in the tree structure and cannot be learned by the recursive network, thus they are set to the mean during the training  and are then optimized with other parameters in the test-time optimization phase. 

For the segmentation experiments where we use the $k$ nearest neighbors algorithm, we set $k=10$ for our results, and $k=3$ for PSegNet results. As for the reconstruction experiments, when using SIREN we need to input the point normals along with the point cloud, for this we use a normal estimation PCA-based method, which fits local planes to each point’s neighborhood.

\section{Additional evaluations}

\subsection{Semantic Segmentation}
We quantitatively evaluate our method's performance on the semantic segmentation task applied to our test sets using standard classification metrics precision, recall, F1 score, and Intersection over Union (IoU). For each class, precision is defined as the ratio of true positives to the sum of true positives and false positives, while recall is the ratio of true positives to the sum of true positives and false negatives. The F1 score is the harmonic mean of precision and recall. Finally, IoU is computed as the ratio of the intersection to the union of the predicted and ground truth point sets for a given class.
 Table~\ref{tab:segmentation} shows the results on all test sets. Our method is robust to noise and performs overall on-par with the strong baselines PlantNet~\cite{LI2022243} and PSegNet~\cite{psegnet2022} on each individual task, while performing all the tasks at once.
\begin{table*}
    {\footnotesize
    \centering
        \begin{tabular}{lcccccccccccccccc}\toprule
        & \multicolumn{4}{c}{Stem} & \multicolumn{4}{c}{Leaf}  & \multicolumn{4}{c}{Petiole}&
        \multicolumn{4}{c}{Cotyledons}
        \\
        \cmidrule(lr){2-5}\cmidrule(lr){6-9} \cmidrule(lr){10-13}\cmidrule(lr){14-17}
                  & Prec.  & Rec. & F1 & IoU  & Prec.  & Rec. & F1 & IoU  & Prec.  & Rec. & F1 & IoU  & Prec.  & Rec. & F1 & IoU \\
       \midrule
       \multicolumn{17}{l}{\textbf{Clean}}\\
       PlantNet    & 85.3 & 88.2 & 86.7 & 76.6 & 98.4 & 99.5 & 98.9 & 97.9 & 82.1 & 71.7 & 76.5 & 62.0 & 98.9 & 94.4 & 96.7 & 93.5 \\
         PSegNet    & 99.1 & 99.3 & 99.2 & 98.5 & 98.4 & 98.9 & 98.6 & 97.3 & 66.6 & 59.6 & 62.9 & 45.9 & 99.7 & 97.0 & 98.6 & 97.3\\
        Ours    & 84.8 & 88.0 & 86.2 & 75.9 & 98.0 & 98.4 & 98.2 & 96.5 & 66.7& 59.4 & 62.2 & 45.9 & 96.7 & 95.8 & 96.2 & 92.7\\
        \midrule
        \multicolumn{17}{l}{\textbf{Noisy}}\\
       PlantNet    & 84.2 & 88.4 & 86.2 & 75.8 & 97.7 & 99.1 & 98.4 & 96.9 & 77.6 & 62 & 69.0 & 52.6 & 96.4 & 93.1 & 94.7 & 89.9 \\
         PSegNet    & 99.3 & 83.6 & 90.7 & 83.1 & 95.9 & 99.9 & 97.9 & 95.9 & 96.2 & 29.6 & 45.3 & 29.3 & 99.6 & 98.1 & 98.9 & 97.7\\
        Ours    & 83.0 & 86.3 & 84.5 & 73.3 & 97.9 & 98.3 & 98.1 & 96.2 & 64.2 & 57.4 & 59.9 & 43.7 & 95.8 & 93.9 & 94.8 & 90.2\\
        \midrule
       \multicolumn{17}{l}{\textbf{Depth maps}}\\
       PlantNet    & 86.2 & 86.8 & 86.5 & 76.2 & 98.5 & 99.2 & 98.8 & 97.7 & 70.2 & 65.6 & 67.8 & 51.3 & 97.4 & 88.7 & 92.9 & 86.7\\
         PSegNet    & 99.1 & 98.9& 99.0 & 98.1 & 96.2 & 99.9 & 98.1 & 96.2 & 98.9 & 54.9 & 70.6 & 54.6 & 99.8 & 98.7 & 99.2 & 98.5\\
        Ours    & 84.6 & 84.2 & 84.0 & 73.2 & 98.4 & 98.6 & 98.5 & 97.0 & 57.9 & 54.1 & 54.5 & 38.8 & 96.8 & 94.8 & 95.7 & 92.0\\
        \bottomrule
        \end{tabular}
    \caption{Comparison to PlantNet~\cite{LI2022243} and PSegNet~\cite{psegnet2022} for semantic segmentation on clean and noisy points test sets using standard classification measures Precision, Recall, F1-Score and Intersection over Union (IoU). All values are percentages.}
    \label{tab:segmentation}
    }
\end{table*}

\subsection{Instance Segmentation}

\begin{figure}
    \centering
    \includegraphics[width=\linewidth]{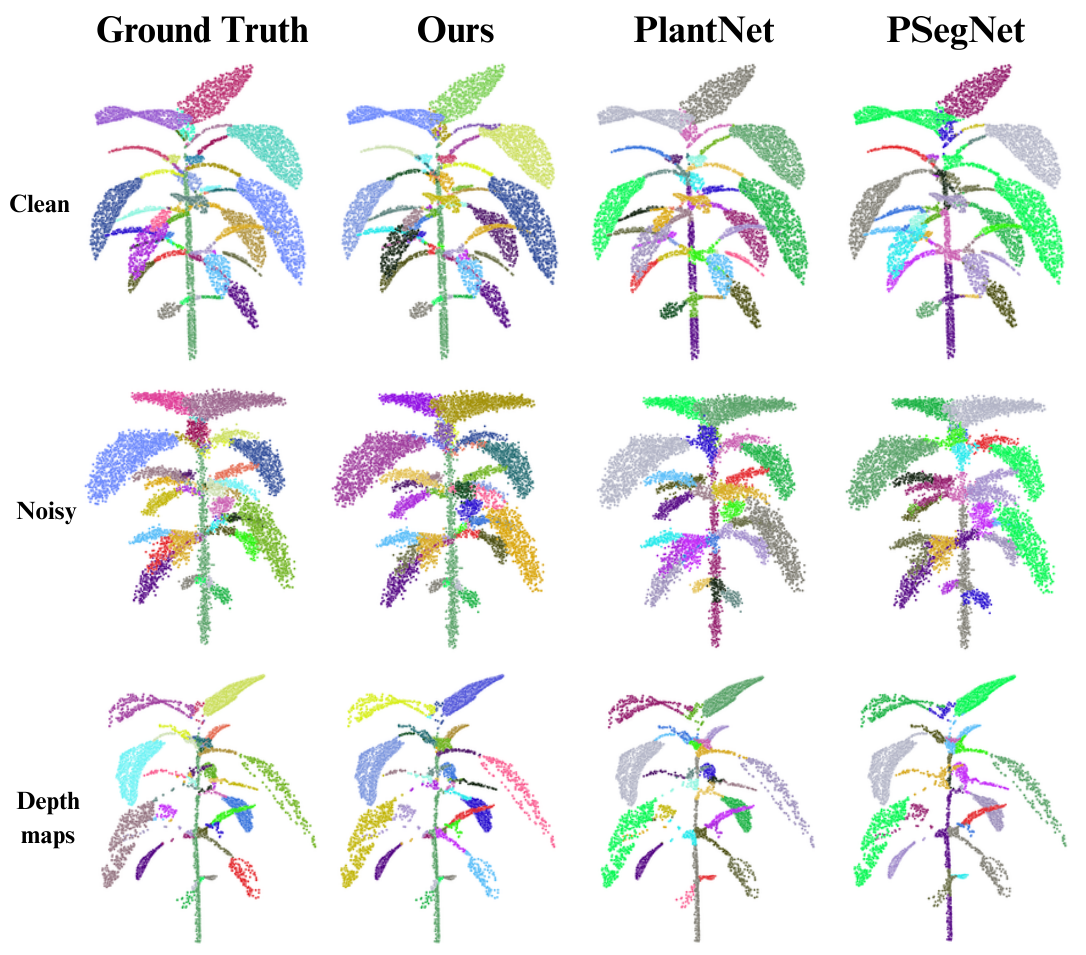}
    \caption{Visual comparison between our method, PlantNet and PSegNet for the instance segmentation task. Different colors are assigned to segmented instances without correspondence to the ground truth.}
    \label{fig:instance}
\end{figure}

We qualitatively compare our method to PlantNet~\cite{LI2022243} and PSegNet~\cite{psegnet2022} for the Instance segmentation task. Fig.~\ref{fig:instance} shows the instance segmentation results on examples from all test sets, where
each individual plant organ is assigned a unique color. Note that the colors are to distinguish the segmented organs without any correspondence between the results and the ground truth. Our method is on-par with the baselines for this task where it is able to segment individual organ instances and is robust to noise and partial data.

\begin{figure*}[h]
    \centering
    \includegraphics[width=\linewidth]{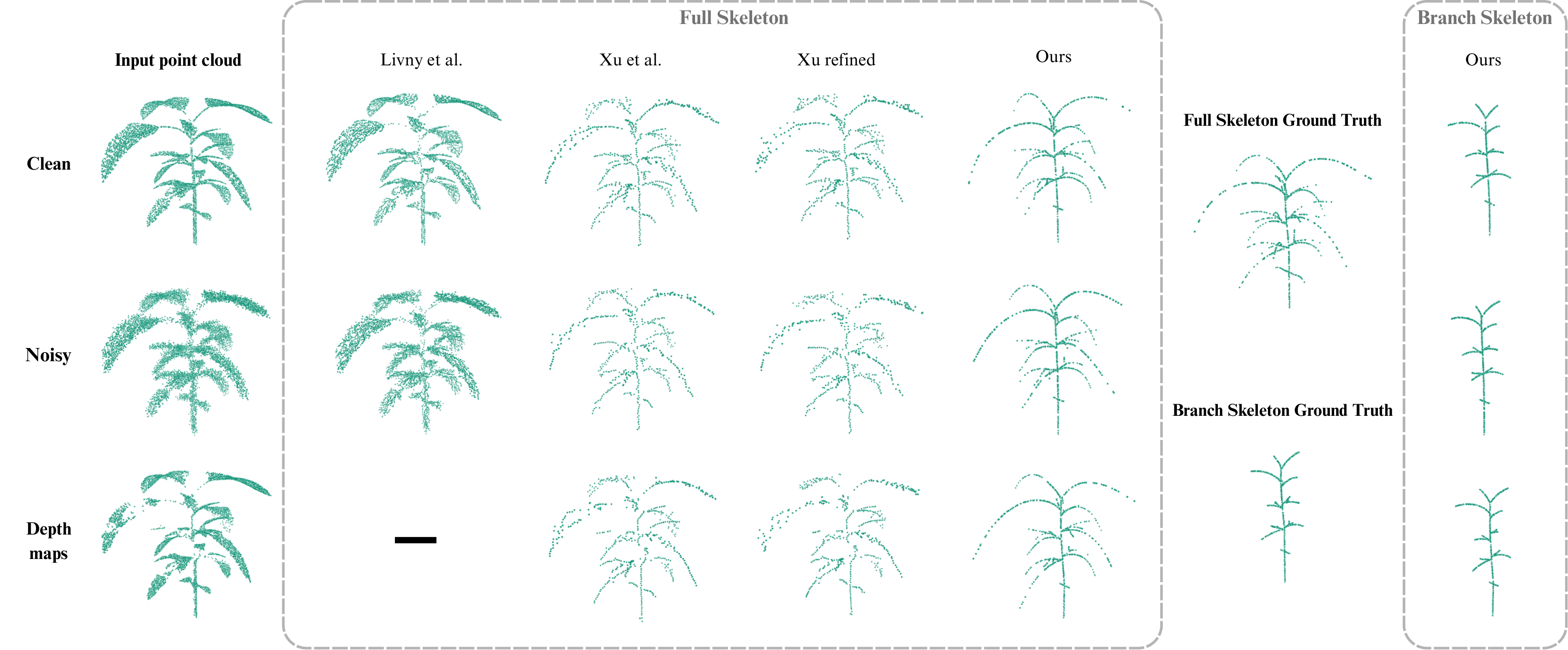}
    \caption{Comparison to Livny~\etal~\cite{livny2010automatic}, Xu~\etal~\cite{10.1145/1289603.1289610},  and Chaudhury~\etal~\cite{Chaudhury2020.02.15.950519} refinement on Xu skeleton for skeletonization. “-” means that the method crashed due to numerical problems. Note that ours is the only method applicable at different scales, and able to output a full skeleton or a branch skeleton, while achieving visually accurate results.}
    \label{fig:skeletons_all}
\end{figure*}

\subsection{Skeletonization}

To support the results shown in Table~\ref{tab:skeletonization} in the main paper on the skeletonization task, we qualitatively compare to the baselines Livny~\etal\cite{livny2010automatic}, Xu~\etal~\cite{10.1145/1289603.1289610} and Chaudhury~\etal~\cite{Chaudhury2020.02.15.950519} on all test sets. Fig.~\ref{fig:skeletons_all} shows the visual qualitative results of all methods on extracting a full skeleton from an input point cloud, with our results of extracting the branch skeleton compared to the ground truth. Our results are close to the ground truth skeletons and are robust to noise and partial input on both scales.

\subsection{Ablation Studies}
Our biologically inspired network uses recursive auto-encoders, which consist of recursive encoder-decoder pairs and classifiers. This idea is implemented using the simplest individual components, namely multi-layer perceptrons with a single hidden layer for the recursive encoders $E$ and decoders $D$, and the classifiers $C_{node}, C_{split}$. Classifiers in recursive neural networks are required as stopping criteria for the recursive decoding and to select the node type; hence, they cannot be removed for ablation. Table~\ref{tab:ablation} provides a neuronal ablation where for each component $C_{node}$, $C_{split}$, and $E\&D$, 75\% of all neurons are randomly deactivated. For each model, we measure the Tree-Edit Distance (TED)~\cite{ferraro-hal-00827474} between input and reconstructed L-Strings of the test set. TED measures the difference between two tree graphs \wrt the number of operations needed to translate from one tree structure to the other, with additional local cost functions corresponding to the difference between the node parameters in our case. Hence, it measures the difference in both topology and shape between plants represented as L-strings. The full model has lowest TED. 

\begin{table}[h]
    \centering
    \begin{tabular}{lcccc}
        \hline
        & Full model & $C_{node}$& $C_{split}$& $E\&D$\\
        \hline
        TED & 0.196 & 0.207 & 0.204 & 0.240 \\
        \hline
    \end{tabular}
    \caption{Table shows an ablation where the full model is compared to models where $75\%$ of all neurons are deactivated in parts of the architecture. For each model, we measure tree edit distance.}
    \label{tab:ablation}
\end{table}

\subsection{Influence of Latent Dimension}
We study the influence of the dimension of the learned latent space $\latentspace$ on the performance of our model. For that, we evaluate the L-string latent representation stage by training the recursive auto-encoders with different latent dimensions $\dimension_{\latentspace}$ and measuring the TED between the input L-strings and the reconstructions. Table~\ref{tab:latent} shows the average reconstruction error in TED for different dimensions, in our implementation we choose $\dimension_{S}$ that is most compact with minimum error.
\begin{table}[ht]
    \centering
    {\footnotesize
    \begin{tabular}{|l|c|}
    \hline
         $\dimension_{\latentspace}$ & Tree Edit Distance \\
        \hline
        32 & 0.227 \\
        \textbf{64} & \textbf{0.196}\\
        128 & 0.197\\
        256 & 0.296\\
        \hline
    \end{tabular}}
    \caption{Analysis of the influence of latent dimension on the performance of the L-string auto-encoder model \wrt the TED.}
    \label{tab:latent}
\end{table}

\subsection{Latent Space Analysis}
\begin{figure}[h]
    \centering
    \includegraphics[width=\linewidth]{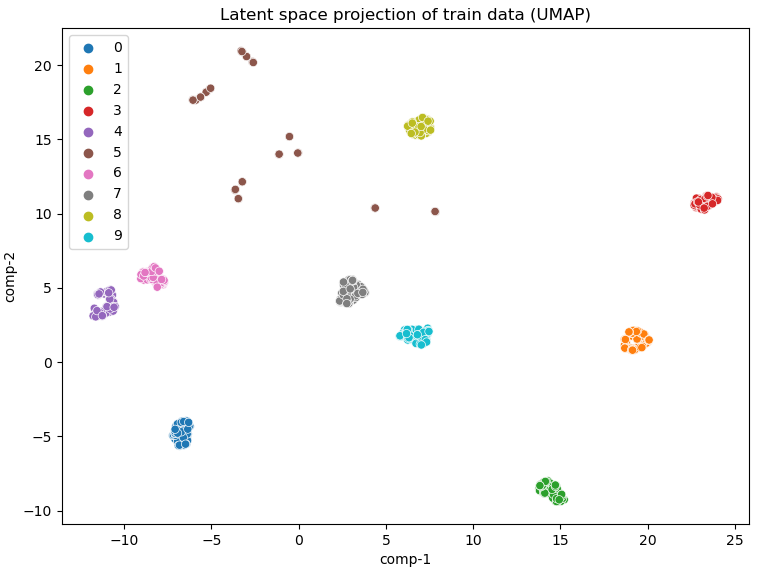}
    \caption{Analysis of the distribution of different tree structures from the training data in the latent space projected in 2D using UMAP. Each color represent a unique plant structure in the training data. Plants sharing the same structure form clusters in $\latentspace$.}
    \label{fig:latent}
\end{figure}

To analyse the data distribution in the learned latent space $\latentspace$, we project the latent points representing plants from the training set in Fig.~\ref{fig:latent} using UMAP~\cite{mcinnes2020umapuniformmanifoldapproximation}, the uniform manifold approximation and projection non-linear technique for dimension reduction with local and global structure preservation. Different plant structures are assigned different colors in Fig.~\ref{fig:latent}, one can notice that our model learns to encode plants sharing the same tree structure close by in $\latentspace$.

\begin{figure}
   \centering
    \includegraphics[width=\linewidth]{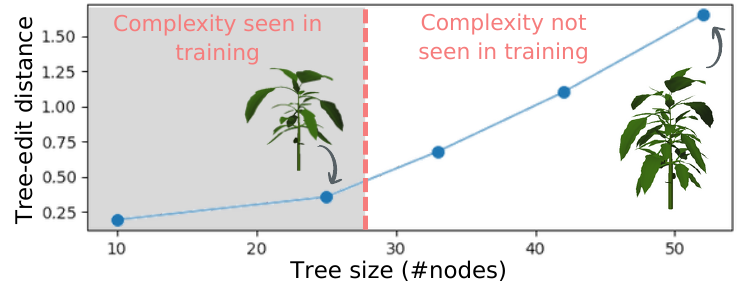}
  \caption{Tree edit distance w.r.t. the development of the plant.}
   \label{fig:complexity}
\end{figure}
\subsection{Generalization to Complex Structures}
To further assess the generalization capacity of our method, we tested how well the trained model generalizes to virtual Chenopodium album in growth stages with a structural complexity higher than that used during training. Fig.~\ref{fig:complexity} shows the TED w.r.t.~the complexity of the plant (using L-String size as a proxy). This indicates that our method generalizes well to virtual plants with higher complexity.

\end{document}